\documentclass[a4]{article}
\usepackage[a4paper]{geometry}
\usepackage{cite}
\usepackage[pdftex]{graphicx}
\usepackage[cmex10]{amsmath}

\usepackage{times}
\usepackage{array}
\usepackage[tight,footnotesize]{subfigure}
\usepackage[font=footnotesize]{subfig}
\usepackage{graphicx}
\usepackage{mathtools}
\usepackage{amsmath}
\usepackage{enumitem}

\usepackage{alltt}
\usepackage{url}

\def\boxx{{\vcenter{\vbox{\hrule height.3pt
          \hbox{\vrule width.3pt height6pt
          \kern6pt\vrule width.3pt}\hrule height.3pt}}\;}}

\def\impos{{\;\vcenter{\hbox{\rule{5mm}{0.2mm}}} \vcenter{\hbox{\rule{1.5mm}{1.5mm}}} \;}}

\def\lrarrow{\leftrightarrow \kern-8pt \rightarrow}

\def\2{\frac{1}{2}}

\def\beq{\begin{eqnarray}}
\def\eeq{\end{eqnarray}}
\def\2{\frac{1}{2}}

\newtheorem{example}{Example}

\def\lrarrow{\leftrightarrow \kern-8pt \rightarrow}

\def\frightarrow{\rightarrow \kern-11pt /~~}
\def\reducesto{\simeq \kern -3pt >}
\def\intersection{\cap}

\begin{document}
\newcommand{\strust}[1]{\stackrel{\tau:#1}{\longrightarrow}}
\newcommand{\trust}[1]{\stackrel{#1}{{\rm\bf ~Trusts~}}}
\newcommand{\promise}[1]{\xrightarrow{#1}}
\newcommand{\revpromise}[1]{\xleftarrow{#1} }
\newcommand{\assoc}[1]{{\xrightharpoondown{#1}} }
\newcommand{\imposition}[1]{\stackrel{#1}{\impos}}
\newcommand{\scopepromise}[2]{\xrightarrow[#2]{#1}}
\newcommand{\handshake}[1]{\xleftrightarrow{#1} \kern-8pt \xrightarrow{} }
\newcommand{\cpromise}[1]{\stackrel{#1}{\frightarrow}}
\newcommand{\policy}{\stackrel{P}{\equiv}}
\newcommand{\field}[1]{\mathbf{#1}}
\newcommand{\bundle}[1]{\stackrel{#1}{\Longrightarrow}}

%\title{The Promise Theoretic Agents and `Attention' in\\Knowledge Representations\\{\large Fast and Slow Variables in Agent Cognition}\Large.}
\title{$\gamma(3,4)$ `Attention' in Cognitive Agents\\\small ~\\{\large Ontology-Free Knowledge Representations\\\large With Promise Theoretic Semantics}\Large.}

\author{Mark Burgess\footnote{Formerly Professor of Network and System Administration, Oslo University College}\\~\\Chitek-i AS, Oslo}
\maketitle

\begin{abstract}
  The semantics and dynamics of `attention' are closely related to
  promise theoretic notions developed for autonomous agents and can
  thus easily be written down in promise framework. In this way one
  may establish a bridge between vectorized Machine Learning and
  Knowledge Graph representations without relying on language models
  implicitly.  Our expectations for knowledge presume a degree of
  statistical stability, i.e.  average invariance under repeated
  observation, or `trust' in the data.  Both learning networks and
  knowledge graph representations can meaningfully coexist to preserve
  different aspects of data.  While vectorized data are useful for
  probabilistic estimation, graphs preserve the intentionality of the
  source even under data fractionation.  Using a Semantic Spacetime
  $\gamma(3,4)$ graph, one avoids complex ontologies in favour of
  classification of features by their roles in semantic processes.
  The latter favours an approach to reasoning under conditions of
  uncertainty. Appropriate attention to causal boundary conditions may
  lead to orders of magnitude compression of data required for such
  context determination, as required in the contexts of autonomous
  robotics, defence deployments, and ad hoc  emergency services.
\end{abstract}

%\tableofcontents

%%%%%%%%%%%%%%%%%%%%%%%%%%%%%%%%%%%%%%%%%%%%%%%%%%%%%%%%%%%%%%%%%%%%%%%%%%%%%%%%%%%%%%%

%%%%%%%%%%%%%%%%%%%%%%%%%%%%%%%%%%%%%%%%%%%%%%%%%%
\section{Introduction} 
%%%%%%%%%%%%%%%%%%%%%%%%%%%%%%%%%%%%%%%%%%%%%%%%%%

The concept of `attention' has come to be associated with the
Transformer Architectures in Deep Learning, as championed by Vaswani
et al in \cite{attention0,attention2,bishop1}, thanks to the successes
of Large Language Models\cite{llm}---though the concept dates back to
the 1990s. In the context of knowledge retrieval, attention refers to
the selective sampling of key-value pairs, through a channel enriched
with semantic annotations called `features'. The goal is to classify
and rank pairs, matching some selection criteria, by measuring their
contextual relevance based on some statistical procedure.

Since the 1990s, attention mechanisms have been used for realtime
context adaptation in agents.  For example, the cognitive maintenance
agent family known as CFEngine\cite{burgessC1,burgessC11}) employs a
closely related lightweight attention method, for the select the
appropriate goals in different environmental contexts.

The purpose of this note is to construct a technology-agnostic
description of attention, using Promise Theory, in order to bridge the
conceptual gaps between `big data' batch machine learning systems and
more focused cognitive agents.  In many ways, the attention matching
concept can be viewed as an implementation of promise theoretic
channel bindings, so the language is easily suited as a generalization
of technology specific vector methods. One may show how to compare and
even integrate vectorized learning models with the ontology-free
$\gamma(3,4)$ representation of knowledge semantic graphs, described
in \cite{burgess_sst2025}. Together, these
representations might be combined for more intentional and accurate reasoning
capabilities.

The outline of the papers is as follows. I recap some of the main
concepts of attention, especially in the language of batch machine
learning, with feature vectors.  These concepts are then rewritten in
promise theoretic notation, which emphasizes inhomogeneous data and
graphical representation. The semantics of space and time are them
introduced via the $\gamma(3,4)$ representation for graphs and we
remark on how this is suited for specialized cognition in more
scalable approaches particularly to autonomous agents.

\section{Elements of attention}

\begin{figure}[ht]
\begin{center}
\includegraphics[width=8.5cm]{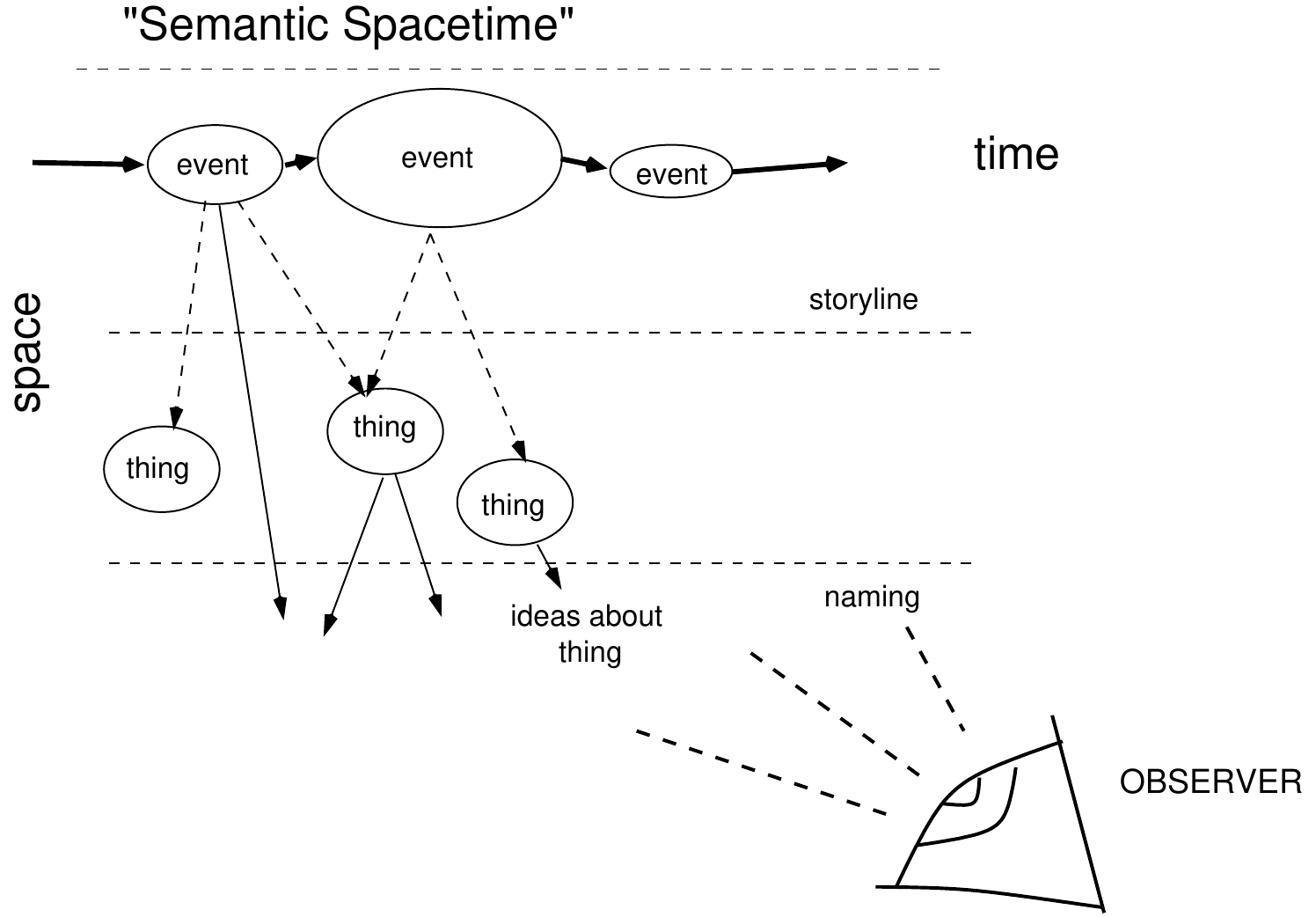}
\caption{\small Semantic Spacetime: an observational process
  collecting data from a causal semantic process has a `knowledge
  graph' that maps the process as a sequence of annotated events,
  forming a fibre bundle topology. We would like to retain this
  information.\label{sst}}
\end{center}
\end{figure}

Consider the observation of some process (both semantics and dynamics) by an
observer agent, which performs selective sampling from its environment (figure \ref{sst}).
The `data' are collected the observational process, and the observer
may be called an `agent', which is simply the embodiment of the
localized process\cite{promisebook}. The semantics of the environment can be
understood in terms of a spacetime model, organized as bundles around fibres of
causal development\cite{spacetime1}.

Localization of the process serves to calibrate both the values and
the meanings of the data received from different external sources at a
single point of definition. This is the key to achieving consistent
semantics, but is also a warning about the intersubjectivity of any
sampling set\cite{observability}.  The end result of sampling a
process is something like a collection of `facts' about the world,
which we wish to record for future reference. However, we may harbour
the prejudice that something we call a `fact' needs to be `objectively
true' in some sense.  The latter is unfortunately unknowable by any
finite process, because the data come from divergent causal origins
that are not arguably comparable.  Moreover, habituation by repetition is not
obviously distinguishable from truth.  One builds up information by
repeated sampling and only later do we validate it with the assimilation of counter-factual
challenges over time\cite{pearl1,pearl3}.  So, instead we shall refer
to such these proto-facts interpreted from data simply as `knowables'
(i.e. things that can be known by an agent, whose truth remains to be
determined), in order to acknowledge their status as more than data,
but perhaps less than fact.

Ultimately, knowledge only refers to something that's actually
`known', i.e.  normalized by repeated experience.  It expresses our
familiarity with concepts, which we summarize as the concept of
`trust'.  Trust has both potential and kinetic
components\cite{burgessdunbarpub}.  The kinetic component, which we
might call `attentiveness' is associated with how much effort is
invested into the selection and sampling of a body of information.
Attention is thus related (dynamically and semantically) to this
kinetic trust.  In turn, the accuracy of observations is related to
the Nyquist sampling rate for the information channel\cite{cover1}. In
contemporary `big data' learning, the body of data is typically a huge
set of pre-collected data for bulk processing. In the realtime
cognitive agent case, data sets would be received more gradually as
time-series\cite{box1} through some sensory interaction, for immediate
response. It's easy for the assumptions and semantics behind these
processes to lost in our eagerness to reach an answer.  Put simply, there is a
deep connection between assessment of data, attentiveness (dynamic and
semantic) and human trust, which has been explored and successfully
modelled in work with
Dunbar\cite{burgessdunbar1,burgessdunbar2,burgessdunbarpub}.

The contemporary term `attention' will, for many readers, have come to
their own attention thanks to reference \cite{attention0}, which
addressed a method for bulk computation across the very large data
sets used to train Artificial Neural Networks (ANN).  For agents at
the edge of the network, such as CFEngine and systems associated with
controlled swarms (such as robots, drones, smart buildings, and
environments with sensory systems and so on), such methods would be
overkill, and one turns to leaner, more energy-efficient approaches to
learning using causal boundary conditions to reduce the data requirements
foe learning.  This points us to distributed systems and agent
autonomy.

In a method dominated by statistical analysis, which washes away
specific contexts, the question of the {\em intentionality} behind a process
we observe remains critically important for modelling
meanings\cite{searle1,burgess_intent}. In general, when we observe
some scenario, there are {\em intentional} themes and {\em ambient}
information that helps to contextualize the main storyline.  The
deconstruction of any scenario into intentional traces and ambient traces is,
in fact, a method that can be applied with the help of knowledge graphs.  Both of
these may be ultimately important as keys for memorization.
$\gamma(3,4)$ is useful because it expressed this intentionality.

In computer science, one most commonly reverts to tropes like Control
Theory, which formalizes `deterministic correction as an axiom' to model system
behaviour as the inevitable outcome of brute force. Promise Theory
refers to this as `imposition'; it is known to be generally ineffective in
practice and somewhat dubious as a concept, except is highly
bounded circumstances.  Physics teaches us that control
is a convenient fiction, at very large scales and under
conditions of sufficient stability. Thus, while on might be able to
get away with ignoring causal intentionality, on some level, for very
large scale computations, with dynamic stability, this will not do for
smaller agents which are common to agents at the interface to their
environments. The development of Promise Theory may be understood
as the effort to incorporate these issues into a simple skeletal framework
for agent interactions.

\section{Knowledge representation and semantics}

There are three common forms of data representation for knowables: the
{\em database} or {\em associative key-value store}, semantic {\em
  knowledge graphs}, and {\em neural networks}, including
contextualized {\em language models}.  Each of these maps key-value
pairs in some sense.  In each case, an element of the representation
may choose to represent a small fragment of information or a bulk
object, like a long text, an image, etc, and equip it with a
searchable key.  The essence of a transformer is to define, determine,
and enrich keys to enable the selection of the most appropriate
values sought be some attending process.

Of these three, knowledge graphs stand out as being substantially
different from the others, because they are generally inhomogeneous
(ad hoc and open ended in data attributes) and maybe used to model all kinds of processes, step by
step. Steps may be spatial, temporal, or conceptual (see figure \ref{sst}). A major
difference between databases and neural network models is that
databases preserve original data, whereas the network learning
fractionates data into vectorized attributes, diluting the original
structure sometimes beyond recognition.  Encodings can preserve
certain aspects of order, but the workarounds chop up and splice
elements rather than preserving original content.

\subsection{Feature-attribute spanning sets}

Stored information may be enriched with annotations, discovered over
time, which help to clarify its semantics. This includes information about
the approximate coarse-grained context in which it was sampled. A set of `features' or semantic
attributes, for each datum, gives it a form of multi-dimensionality in
a conceptual `space' of ideas. If such annotations are homogeneous for
all data points, one can form a rectangular matrix of {\em data}
$\times$ {\em annotations} (usually denoted $X$), so that the annotations become simply part
of the data in a regular schema. In other cases, annotations may be
highly inhomogeneous, in which case a graph representation is more
favourable since this makes no assumptions are about the completeness
or homogeneity of data.

\begin{figure}[ht]
\begin{center}
\includegraphics[width=7cm]{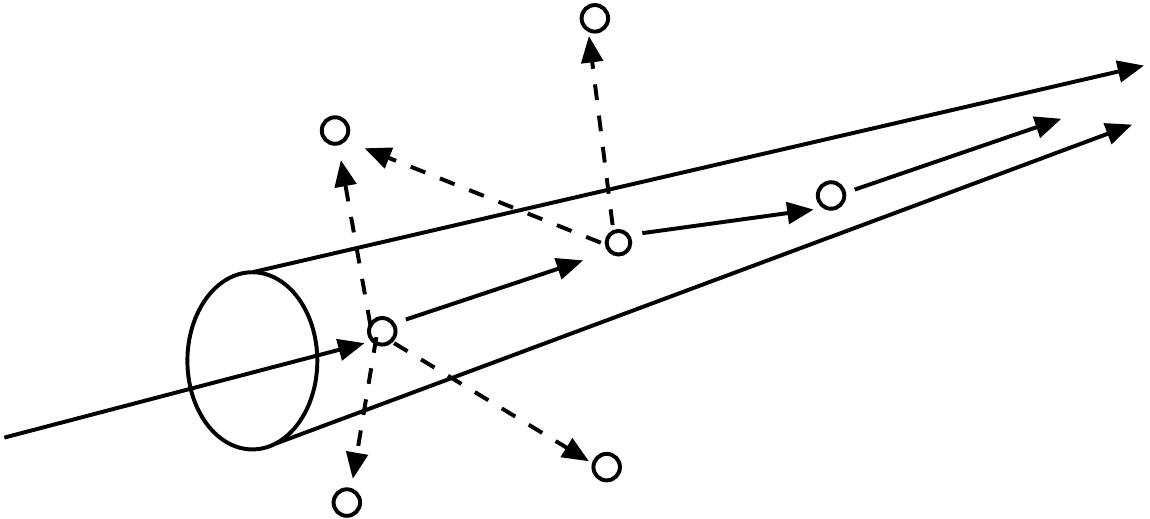}
\caption{\small Cognitive data have the structure of a semantic spacetime, with attributes orbiting
around the events joined into finite length causal spines. These paths generally terminate
due to contextual semantics of real world processes, whereas raw data may continue as an endless timeseries.\label{fibre}}
\end{center}
\end{figure}

In either case, semantically rich processes have the structure of a spine
of causally-related central events, enriched by orbiting annotations of each event
in the form of attributes orthogonal to the central node (like a topological fibre bundle),
though certain attributes may orbit more than one centre in a graph, which tells us that
a vector representation isn't semantically appropriate (see figure \ref{fibre}).

One often speaks of dimensions in a geometrical way
(not just a set theoretic way), by imagining a picture of data in a
$D$ dimensional vector space. This inference of a geometry is speculative and rooted
in cultural norms, though it has led to some interesting applications\cite{word2vec}.
Each dimension represents a linearly
independent attribute. For instance, in human vision, one can invent an
approximately accurate 3 dimensional space for colours spanned by red,
green, and blue light intensities (at least for those who are not
colour blind).  Given a spanning set (basis) of primitive sets 
$\phi_d = (R,G,B)$, for $d = 1,2,\ldots,D$, we can think of these as basis
vectors or sets as we please. 

Geometry is always an intuitive choice, but it also overestimates the
amount of freedom available to feature classification.  Thinking about
colours, once again, there is no infinite resolution in the blending
of colours accessible to any human or non-human eye.  The finite
number of bodily cells and finite signalling capabilities will always
project the distinguishable values into a finite set.

For different representations, basis functions $\phi_d$ may take the
form of data types, vectors in a vector space, matrices or other
matroid basis, graphs (products of nodes and links), general sets, and
so on.  They correspond to a set of independent sensory inputs, which may or may
not overlap in their detection capabilities.
Each representation is suited to certain operational uses,
with varying degrees of convenience.  Using Promise Theory, we'll see
below how all these can be represented as autonomous agents, with
simple set theoretic languages, that voluntarily forego part of their
autonomy when interacting with other agents, thus yielding a single
methodology for all representations.

\subsection{The promise theoretic interpretation}

Promise Theory was introduced in order to deal with the failures of
logical type reasoning in computer
science\cite{burgessDSOM2005,promisebook}.  It was derived originally
from extensive practical experiences in the use of the CFEngine family
of agents\cite{burgessC1}, which have been widely deployed to manage
datacentres around the world since the mid 1990s. In particular, it
incorporates the concepts of autonomous behaviour, with commination
over a partially reliable information channel from Shannon's Theory Of
Communication\cite{shannon1,cover1}, and extends it with semantics for
a more complete picture of informatic behaviour. General results for
Promise Theory apply to any computational mechanism, thus it unifies
across implementations and exposes potential flaws and uncertainties
in models.  It forms a lingua franca for analyzing intentional agents
and their causal processes.

CFEngine, as a system of autonomous software agents, introduced
mechanisms for quickly attending to sensory signals and classifying
decision making trajectories from a mixture of data from its
environment and the system's state of repair.  This is the same
approach used in large token models today, except for being at much
smaller data dimensionality, which implies lower statistical certainty
and thus the need for alternative approaches to achieving stability.

Promise Theory describes all phenomena in terms of causally independent,
or autonomous entities which it also calls agents.  An `agent', in the
sense of Promise Theory, is therefore any active location with its own
resources and independent causality. Agents are separable process
elements, which stand alone and keep promises as a service to other
agents.  This is general enough to include more general definitions of
agents, such as software processes and robotic machinery.

By making promises to each other, promise theoretic agents select a
causal direction, and may therefore be said to `intend' to forego part
of their autonomy and signal their {\em intent} to behave in a predictable
way.  Intentionality thus forms a significant element in promise representations.
Agents thus assume the roles of promiser and promisee.
Further, they may be composed to form superagents, or decomposed to
identify subagents, should their interior details become observable.
In general agents are `black boxes' which signal their intentions as
formalized `promises'. Thus, as a kind of universal language of
intent, promises and agents may be used to model any scenario: real or
virtual.

Since promises describes a network of relationships between agents,
they lead to an immediate graph representation for both physical and virtual
processes. The relationship between promises and knowledge graph
representations was described in \cite{spacetime1}. The promise graph contains
more information than a basic graph $\Gamma(N,L)$ of nodes $N$ and links $L$. Below, we'll
explain how to unify a promise graph with probabilistic selection of
attributes used un vector machine learning (as used in Large Language
Models\cite{duda1,bishop1}). This is important, because graphs are
more obviously deterministic in retaining causal relationships, and
thus simplify the encoding of intentional information lost by
fractionation in vectorized models. 

A known obstacle to using graphs for knowledge representations is the
reliance on ad hoc ontologies to inject missing structure, in traditional modelling approaches based on
types. The Semantic Spacetime model overcomes this problem, however,
by replacing rigid logical typing with causal discrimination and
collective coarse graining.

Following the conventions of Promise Theory\cite{promisebook}, every
active and passive element in a system is considered an agent
generically labelled $A_i$, for $i=1,2,\ldots$. The agents' attributes
and promises are to be specified.  Each agent has some interior
resources, which are unobservable to other outside agents, and acts
autonomously. Agents are typically inhomogeneous both through their
interior capabilities, and in virtue of the promises they advertise
outwardly. A promise is a signal of an agent's intent. Agents can only
make promises about their own behaviour, as is the meaning of
autonomy.

A network of promises thus represents a directed multi-typed graph.
Any graph structure can always be represented by atomic triplets of the form $(N_S,L,N_R)$,
where $N_S$ and $N_R$ are graph nodes (also called vertices), and $L$ is a (directed)
link (also called an edge) pointing from $N_S$ to $N_R$. A graph is a collection of 
many such directed triplets, and a promise representation of the graph makes every node into an agent,
and every link the result of two promises:
\beq
\pi_S : S &\promise{+b_S}& R\nonumber\\
\pi_R : R &\promise{-b_R}& S,\label{pt}
\eeq
where $b_S$ is the body of the intended offer by $S$ ($+$ implies an offer) and $b_R$ is
the body of the intent to receive ($-$ implies acceptance) by $R$.
\begin{figure}[ht]
\begin{center}
\includegraphics[width=6.5cm]{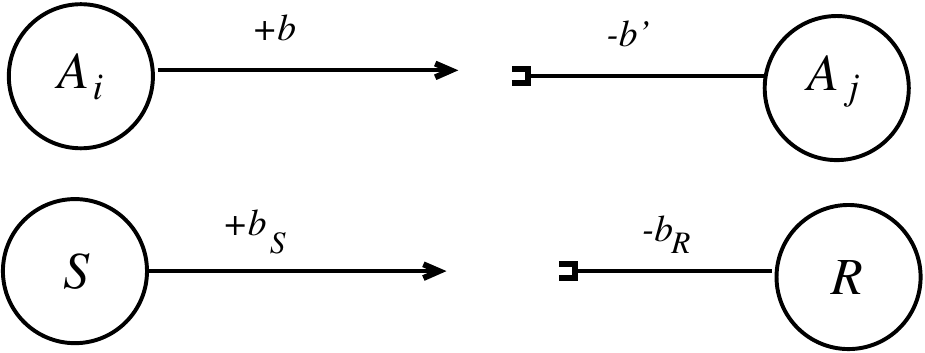}
\caption{\small A schematic promise graph notation for equations (\ref{pt}). Note that two promises are
required to make a single conventional graph-theoretic link.}
\end{center}
\end{figure}

Notice that the autonomy of agents $S$ and $R$ means that both agents need to keep a promise
in order for there to be communicated information. The intended information
travels in one direction, from $S$ to $R$, although the prior signalling of intent
involves a handshake in both directions. The extent of what is communicated 
is bounded by $b_S \intersection b_R$, where $b_S$ and $b_R$ are set valued.
This overlap forms the basis of both the type semantics $\tau$, intended context $\chi$,
and numerical `weight' (e.g. probability, potential, trust, etc) of a traditional graph link:
$L(\tau,\chi,w)$.

Lastly, because each agent is an autonomous dynamic process, with
potentially interactions in-between, each agent also samples signals
and data from other agents itself. The sampling process involves
measurement, and thus each agent must be considered to possess a
function which acts as an `assessment' capability to determine whether
another agent's promise has been kept or not.  The assessment is
autonomous, so one can't assume that data will be captured and passed
on with fidelity and accuracy.  The ability of an agent to interpret
promised data depends on its own sampling rate relative to the rate of
change of the promised information, as well as the resolution, and is
written $\alpha_A(\pi)$.  The ability and `willingness' or {\em intent} to pass on the
information accurately is a separate issue.

Our goal now is to apply these limitations to data and machine learning.

\subsection{Chain rule for propagation of promised information in learning optimization}

In machine learning networks, like ANN, potentially large numbers of agents
collaborate in the representing a data pipeline whose overall promise is to
rank or classify the inputs according to a policy.
Agents make promises one-by-one, but the movement of information
through a network of agents is called virtual motion of the third
kind\cite{virtualmotion1}, and is subject to potential distortions as
a consequence of the Intermediate Agent law\cite{promisebook}.  In
machine learning parlance, this is modelled implicitly by
computational transformations, which are treated as exact
deterministic relations. That which is sent by one layer of a neural
network is that which is received by the next, without question.  The
process may therefore be represented as algebraic matrix
multiplication\cite{duda1,bishop1}.

Promise theoretically, we refer to a source agent $S$ as being `upstream' and a receiving agent $R$ as being `downstream'
of a flow of information. An intermediate agent $I$ can make a conditional promise to relay
information $x$ from $S$ to $R$, which makes its relayed information dependent on what it receives.
\beq
S \promise{+x_S} I \promise{x_I | x_S} R
\eeq
and
\beq
I &\promise{-x_S}& S\\
R &\promise{-x_I}& I
\eeq
Moreover, it will only relay the information if it assesses that the promise
from upstream was kept, i.e. that it has received information it judges (autonomously)
to be intentionally sent from $S$.
If one defines a promise matrix over all agents $A_i,A_j$, each of which
may be in the role $S$ or $R$ for different promises\cite{spacetime2,virtualmotion1},
by the Hadamard product:
\beq
\Pi^{(\pm)}_{ij} = \Pi^{(+)}_{ij}(A_i)\,\Pi^{(-)}_{ij}(A_j),
\eeq
we can separate the contributions to this serial dependency agent by agent.

Let's identify notational representations and assumptions.
In a neural network (ANN) representation, the 
nodes $s_i$ are agents in a network, arranged in layers $\ell$ with results written conventionally:
\beq
\vec y^{(\ell)} =  \vec s^{(\ell)}
\eeq
This layer promises to forward information to the next layer:
\beq
\vec y^{(1)} &=& \vec f\left( \sum_i  w_i^{(1)} s_i^{(0)} + r^{(1)} \right),\\
\vec y^{(2)} &=& \vec f\left(  \sum_j  w_j^{(2)} f_j \left(\sum_i w_i^{(1)} s_i^{(0)} + r^{(1)}\right) +r_j^{(2)} \right),\\
\eeq
and so on, iteratively, were $f()$ is the `neural' activation
function, which is conventionally assumed to be homogeneous throughout
the network. The autonomous contributions $r_i$ can be absorbed into
the definition of the $f_i()$, This forms a piecewise synthesis
of the function envelope. The key structural feature is that each layer
builds on the linear aggregation of upstream values $\sum_i w_is_i$
arriving in the pipeline, and that non-linearity is built by recursively
activating function $f()$ to perform envelope shaping.

When using this in ANN optimization, it's conventional to use a self-attention continuum embedding
by replacing source values $s_i$ with a `softmax' surrogate value, where
\beq
\text{softmax}(w_i) = \frac{e^{\beta w_i}}{\sum_i e^{\beta w_i}}  \le 1.
\eeq
This policy operation tends to enhance the maximal values, sharpening or reducing the Shannon entropy
of the distribution. However, it's expensive to compute, since normalizations are needed for 
values over each layer, this requires having every agent send its value to
be summed at a single point of calibration. One could introduce a single
agent per layer, but the obvious recipients would be the agents of the next downstream layer, each
of which must then independently compute the sum. This is an expensive inefficiency whose accuracy
has unclear implications for the results. The embedding aspect of this
could be maintained with an inverted minimization:
\beq
\frac{e^{\beta w_i}}{\sum_i e^{\beta w_i}} \rightarrow (1 - e^{-\beta w_i}) \le 1, 
\eeq
which requires normalization but does not result in equal sharpness (since it is now inverted maximum 
rather than entropy). However, since the policy is ad hoc, adjusting the
argument of the exponential to find a satisfactory alternative could be assessed. One advantage of this
approach is that realtime updates can be admitted without massive recomputation, by
forming convex updates $\Delta w$ with some policy $c < 1$:
\beq
w_{t+1} = c\,w_t + (c-1)\,\Delta w.
\eeq
This leads to a geometric series in $c$ which forgets older values in a predictable manner---a
feature which is highly suitable for realtime agents like robots.
 
Written as a calculation like this, there is no question about timing or reliability of the channel.
It is simply a time-honoured matrix computation. A master process governs the coordination.
We now wish to treat these as being part of a noisy channel by writing the above in promise
notation, which embodies that assumptions of causal independence and Shannon information.
In promise form, this looks something like the following\cite{spacetime2,virtualmotion1}:
\beq
\prod_i \left( \Pi^{(\pm)}_{si} \Pi^{(\pm)}_{ir}\right)
\eeq
and the transmission bandwidth $b_{ij}= b_i \cap b_j$ can be supplemented with
the assessment function $\alpha_i(\pi)$ for each agent $A_i$ and promise channel $\pi^{(+)}\pi^{(-)}$.
The assessment function plays a complementary role to that of the activation function in an ANN array.
The activation function belongs in the conditional chaining:
\beq
A_{\ell} \promise{+f(b)\, |\, b} A_{\ell+1},
\eeq
for appropriately promised function $f()$. Thus, combining everything, we have:
\beq
Y = \alpha_R\left( \prod_i \left( \Pi^{(\pm)}_{si} \Pi^{(\pm)}_{ir}\right)\right).
\eeq

\subsection{Semantics as boundary conditions over time and space}

One of the weaknesses of brute force machine learning, lies in trying
to handle everything by a single batch method fed to a variety of
downstream neural network models. Causal information is stronger
upstream.  Statistical knowledge relies on increasing certainty
through bulk repeated sampling.  For lean cognitive agents, repeated
sampling is a pseudo-continuous process. If the source signal is
invariant (constant), then repetition quickly eliminates value
uncertainties associated with downstream measurement (by the Nyquist
law\cite{cover1}). If the signal is varying at some rate
$R$, then the repetition at a rate $\ge 2R$ is required for capturing
a representative pattern.  In the ideal case, patterns will repeat
over epicycles so that we must align samples into corresponding
measurements (see figure \ref{weekly}).

Learning may be either supervised (pre-labelled) or unsupervised
(unlabelled), or a compromise between the two which involves guiding
collection by sampling within the bounds of an assumed pattern. A
simple example of this is the way in which traffic patterns (real and
virtual) are driven by the sociology of the human working week.
Traffic volumes on services: for instance, traffic typically peaks on
Monday or Tuesday, then tails off towards Friday, and is lower on
Saturday and Sunday. This pattern repeated reliably over the week,
because the weekly period is the strongest pattern. There is also a
pattern of variation over the course of a day, but---because there are
much larger differences between the values sampled across different
days, including timezone variations---a daily model has much higher
variance than a weekly model. In physics, one might call the
statistical variation a non-equilibrium process\cite{burgessIJMPC}.
An initially non-Gaussian process may be transformed into an approximately
Gaussian one by reparameterizing the data.

There is a local equilibrium on a timescale of hours, over which one
can estimate a meaningful average behaviour, subject to some noise. If
one divides the local mean by the local standard deviation (scale of
noise), the result follows a predictable (indeed calculable) pattern.
This is how data can be fitted to a calculated model.  A simple
intentional parameterization of time as an offset to a periodic
process (clock arithmetic) in order to respect the known boundary conditions,
reveals this pattern quickly, but a search using linear statistics
would be more spurious (see appendix).  
\beq 
t = nP + \tau 
\eeq 
This underlines the difference between knowledge representation and
statistical analysis: in representation one is encoding knowledge
intentionally, not fishing in a barrel of data. Thus, the role of
intent is not to be despised. It is a surrogate for learning supervision.

In a similar way, the Transformer Model\cite{bishop1} takes carefully
extracted feature annotations and attempts to scale away dynamic
variation to give a fair prediction.  Over any timescale, we can
rescale by an activity level, or normalize a sample size, by an `energy
of attention' to normalize it as a dimensionless variable.

The key point is that uncalibrated data are generally meaningless. Meaning is
calibrated by working with dimensionless variables, relative to scales that
represent variational features of the phenomenon one is trying to learn.
Promise Theory further relates this sampling process to the agent's own
cognitive experience, and one can define its `kinetic trust' based on
its allocation of resources to sampling a source.  Potential trust is
gauged by the assessment of whether the source keeps its promise or
not\cite{burgessdunbar2,burgessdunbarpub}. The promise model shows us
how to build a notion of trust and trustworthiness purely from the
axioms of autonomous behaviour.

Certain patterns can be identified more cleanly than others. By
minimizing the variability around the intended target, as a fixed
point of the dynamic map\cite{chaosbook}, one regulates the strongest
pattern and identifies this with the intent of the agent.  In the
absence of an intended target, one has only the spurious mean of the data to
minimize around \cite{couchDSOM2003,burgessC11} to an intended
outcome.

\section{Transformers and neural network data}

In Deep Learning\cite{bishop1}, data models are trained by
batch processing rather than by continuous cognitive updating.  The
Transformer Model\cite{bishop1} is a statistical, feature matrix
formulation of key-value associations, which point to a context
dependent generalization of key-value pairs. In a traditional
database, key-value pairs are one-to-one associations; however, in a
network of contexts, a search may result in several different answers
depending on some control parameters. The control parameters may too
many and complex to embed inside a static schema, as in a
traditional database---though some such representations do exist
for basic text searches.

In more complex scenarios Artificial Neural Networks (ANN) are used to
classify keys formed from many parameters, spanned by $\phi_d$, in
order to compress them into a `reduced dimension' model.  This starts
from a premise of an open ended associative pattern, which is then
refined back to `one-to-one'-ness, using use a graph theoretic
structure to generalize and narrow searches based on likely context.

Using vectors to represent semantics can lead to spurious or illusory
identifications based on statistical artefacts.  For example, a data
object might represent languages, with values: English, French,
Chinese, Spanish, etc, or items mail-ordered from a warehouse: shoes,
kettle, tent. There is no vector transformation that can translate or
rotate English into French, in spite of some common dependencies; nor
is there a relationship between shoes and kettle. There is no location
in a meaningful space, which is half way between a shoe and a tent.
One can always embed a set into a Euclidean space, but this is a
representational faux pas in general to do so.

Attention-based decision scenarios can be modelled as an abstract
agent with a query $Q$, browsing a key-value store. On is trying to
find the best (or perhaps merely a good) fit between the intentional
query $Q$ and some key $K$, and to return the associated value $V$,
where all of the triplets $(Q,K,V)$ are sets projected through a
single common dependency, which we might call `the data'. 

\begin{itemize}
\item $Q$, Query representation: a set of (-) promises indicating acceptable responses, expressed as the spanning basis $\phi_d$.
\item $K$, Key representation: a set of candidate (+) promises pointing to values, classified by $\phi_d$.
\item $V$, Value data: invariant source sample corresponding to $Q \cap K$.
\end{itemize}
Attention applies twice: first on the original source training data, and then again when attempting to
recall, match, or predict outcomes. It therefore forms a semantic bridge between phenomena and knowledge representations.

\begin{example}
In batch machine learning, attention is implemented by the transformer matrix architecture\cite{attention0}.
The data are derived by atomizing training sets, and are
pre-processed in batch to optimize a statistical matching algorithm. 
The attention to semantics is something of an afterthought, implemented
as a separation into different `features' and `heads', which annotate
the data in the form of high dimensional vectors.
\end{example}
\begin{example}
For cognitive agents, attention is implemented by defining `smart sensors', which
are optimized to observe and classify particular aspects of the environment, e.g.
vision, sound, text, facial recognition, etc. In CFEngine, the agents of the promise
graph may be identified as follows: the data source $A_S$ yielding $X$ is the set of host computers, their kernel
parameters and filesystems. The context span $A_F$ is a set of `hard classes' based on a study
of universal characteristics\cite{burgessC8} and the cf-monitord agent, along with an extensible
set of `soft classes' from local adaptive environments; these may be denoted $\chi$. The managing agents $A_R$ and $A_T$
are embedded in the cf-agent agent. Finally, the $A_Q, A_K$ transformations of $X$ project one to one
as the hard and soft classes, which take values in the range $\{$ `true', `unknown'$\}$, and the
values $A_V$ are selected by cf-agent from a contextual policy set $\{(\pi^+,\chi)\}$.
\end{example}
\begin{example}
In knowledge graphs, attention results in an inhomogeneous encoding.
Traditional graphs based in RDF/OWL use query languages and attempt to span
features using logical type definitions\cite{rdf}. In $\gamma(3,4)$ 
Semantic Spacetime graphs\cite{spacetime3} (as used in
\cite{SSTorytime}), classification of key-value associations is based on both boundary conditions and 
phenomena, with attributes encoded either by selective discrimination or coarse graining aggregation, 
based on the naming of relationships. The relationship names play the role
of specific types, but these are mainly treated as a projective geometry
with only four dimensions. Context $\chi$, classified by the
sensory system, is built into links $L(\chi,\phi_d)$.
\end{example}

A significant different between data ingested for implicit training
versus data entered intentionally is the loss of intentionality.  When
one makes a note of something specific, the purpose of the note is
highly intentional and the meaning of the information is highly
focused, often in terms of key words that are very close to a spanning
set $\phi_d$.  By contrast, when one reads a book---which is only partly
about issues understood by the sampling set---there may be several purposes embedded in the
free text and the bulk text mixes several raw concepts
together. Although one can always span a text with some set $\phi_d$,
the purity of the decomposition depends on how closely the basis aligns with
the source data. For this reason, machine learning optimizations often
look for eigenvector representations of the input data---however, an eigenvector
basis in the training data may be very unlike a spanning set belonging to
a user who wants to query the data. Finding a plausible mapping between
these spanning sets is the central challenge of attention modelling. 

\subsection{Attention for homogeneous feature vectors}

Let's recap the discussion in \cite{attention0,attention2} for reference. Data
`feature vectors' are $D$ dimensional non-Euclidean sets, in which there are
no a priori relations between the components along different
dimensions. In practice, the dimensional attributes many not be fully
independent, like the different member elements in a database schema
or a programming object, and this can induce a kind of partial
geometry by linear dependence. This could be considered an artefact of
the representation, but it's largely unavoidable due to the nature of
real world processes.

I'll use the nomenclature `interior' for the $D$ data attributes belonging to each $x^{(n)}$
and `exterior' for the collection of many $x^{(n)}$, since the picture is that each $x^{(n)}$
is wrapped in some kind of conceptual boundary that we'll later associated with agents in Promise Theory.
We might also loosely refer to these as interior and exterior spaces, in a colloquial sense---but, we
must be clear that these are not vector spaces.

We start with $N$ sets of homogenous data $\vec x^{(n)} =
(x_1,x_2,\ldots,x_D)$, for some $n$, which we write as vectors for
notational convenience only. This vector is imagined (whimsically) to
belong to `feature space'. This means that each of the columns $x_d$ are assumed
to have identical interpretations or `data semantics'.

If one has $N$ such sets of data, all with the uniform interior dimension $D$, they can be
arranged into a non-square matrix $X$, whose rows are data items and whose columns are the
features for each $\phi_d$. In general, the data will not be numerical, but one can always seek an
artificial numerical encoding, albeit with limited semantic interpretability.

\beq
X = \left( 
\begin{array}{c}
\vec x^{(1)}\\
\vec x^{(2)}\\
\ldots\\
\vec x^{(N)}\\
\end{array}
\right)
=
\left(
\begin{array}{cccc}
x^{(1)}_1 & x^{(1)}_2 & \ldots & x^{(1)_D}\\
x^{(2)}_1 & x^{(2)}_2 & \ldots & x^{(2)_D}\\
\ldots\\
x^{(N)}_1 & x^{(N)}_2 & \ldots & x^{(N)_D}\\
\end{array}
\right) 
\eeq 
This matrix is not a geometrical representation, a
priori. It resembles more closely a graph {\em incidence matrix} in which the
features are links. It represents a batch rather than a geometry.
Geometry can only be returned to the representation through their
original process semantics. This is the goal of the Semantic Spacetime
model, developed from Promise Theory.  The interpretation of this
matrix is thus that of a batch of data rather than a transformation.
However, we can consider transformations of this batch.

At this point, we have invented a vector space that didn't exist in the data themselves. Just as a best-fit line
might not actually pass through any real points, it represents a kind of average that we hope represents
a reality, because we can't really be sure the original data represented reality either. The point
of data science is to deal with uncertainties and informed speculations: we have left the realm of
real things in favour of models. 

As a non-square $N\times D$ dimensional matrix, we can left-operate on the source data $X$ with various
$n \times N$ matrices (acting on different exterior samples), or right-operate with $D \times d$ matrices
that act on the different interior attributes, where $n$ and $d$
are to be decided. The transformer method uses right multiplications by an $D\times D$ operator
to perform an automorphism of the batch collection, taking this three different causal
directions, called query $Q$, key $K$, and value $V$\cite{bishop1}:
\beq
X \rightarrow X\,W^{(Q)} &=& Q\\
X \rightarrow X\,W^{(K)} &=& K\\
X \rightarrow X\,W^{(V)} &=& V
\eeq

One assumes that multiple data `vectors' collected are somewhat related. Otherwise, only
madness lies in collecting them together in this way. The genuine conundrum is: how
related do they need to be? Common assumptions and unified intent are the unspoken 
criteria.

The use of matrices in a geometric space is significant because it
compresses the possible transformations into groups, like the
Poincar\'e group, involving a finite alphabet of algebraic elements,
such as translations and rotations. Shapes are meaningful and are
preserved.  For arbitrarily aggregated data sets, there is no such
group constraint, and therefore no compression of the alphabet of
possible transformations, in general. One can imagine a practically
infinite set of matrices representing different transformations, which
don't preserve shapes, because the semantics are different. This is not to say that
there aren't constraints in practice. Indeed, the programme of machine learning
and pattern recognition is, in broad strokes, to identify such shapes from
the data without prior knowledge. The issue is then: in those cases where we do know
something in advance, how can we preserve that information rather than trying to
guess it unnecessarily?

\subsection{Agent-Promise representation}

We now set aside the background case of using agents in a learning fabric, and
turn to the computation of attention based on a set of data for arbitrary agents. 
We begin by sketching the topology of the transformer attention process described
in \cite{attention0}, which is standard for batch processing.
Consider agents that play roles in the process data retrieval process.
\begin{itemize}
\item The end user $A_U$.
\item The transformer $A_T$.
\item The data source $A_S$.
\item The feature policy source $A_F$.
\item The data curator and feature extractor $A_R$.
\end{itemize}

\begin{figure}[ht]
\begin{center}
\includegraphics[width=8.5cm]{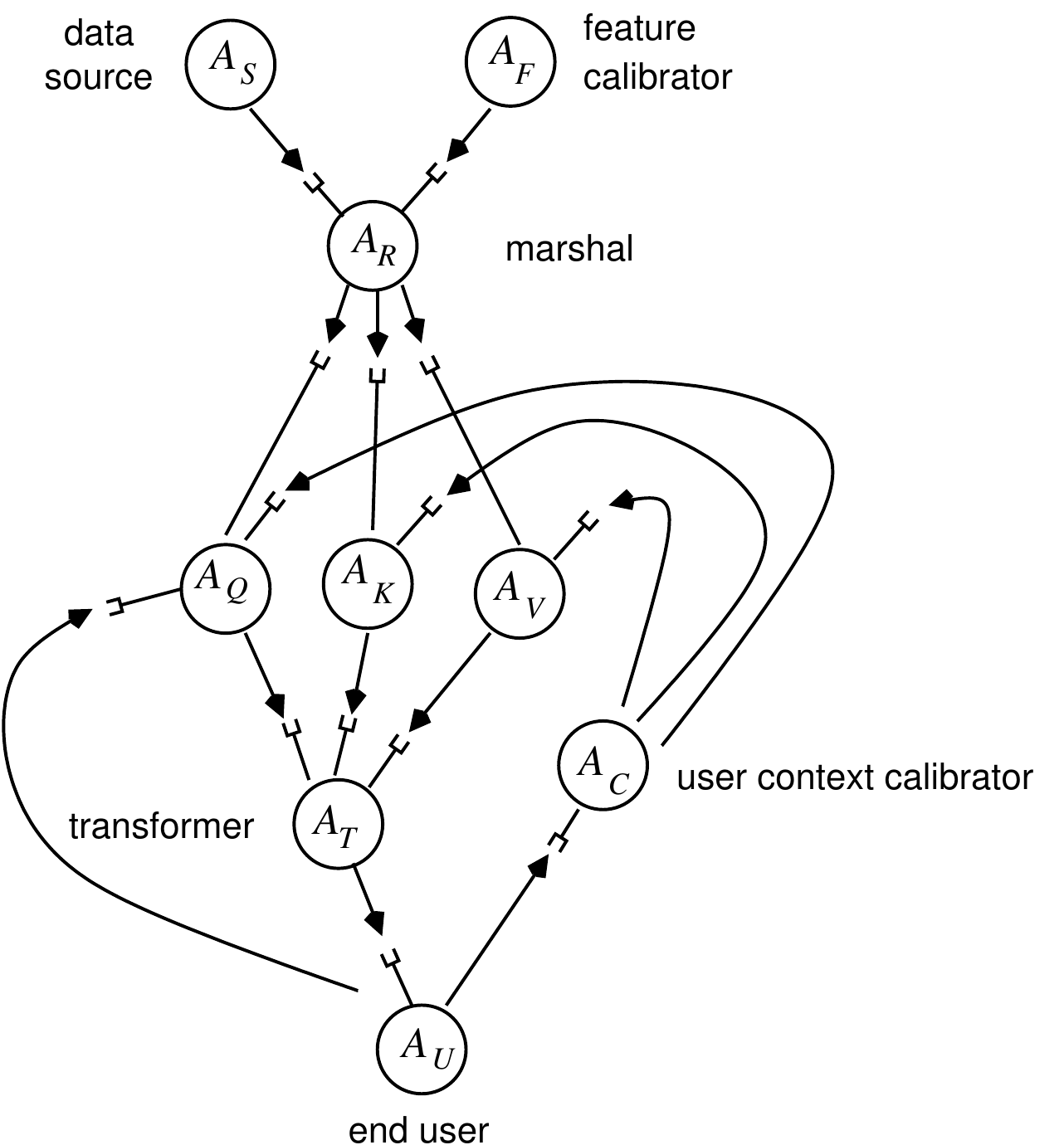}
\caption{\small The transformer architecture expressed as autonomous agents.
One sees the topology dictates the extent to which agents forego their autonomy
in order to coordinate. The tighter the promise bonds, the more rigid the cooperation.\label{transformer}}
\end{center}
\end{figure}

All quantities here are set valued, without any particular structure.
One can say that the precise interior representation is a private interior
property of the agents.
A source agent $A_S$ offers original data $X_S$ to a receiver/curator agent $A_R$,
which selects and accepts some version of it\footnote{It is conventional to label
sender and receiver roles as $A_S$ and $A_R$ in order to emphasize their roles
as ends of a Shannon information channel, with bandwidth $b_S \intersection b_R$.}.
\beq
A_S &\promise{+X_S}& A_R\\
A_R &\promise{-X_R}& A_S.
\eeq
The second and third sources are a feature agent $A_F$, which
promises a list of $D_\phi$ feature attributes $\phi_d$ ($d=1,2,\ldots, D_hi$) used
to span the possible attributes of the data, and a source agent for relevant context $A_C$ (see later below):
\beq
A_F &\promise{+ \{\phi_F\}}& A_R\\
A_R &\promise{- \{\phi_R\}}& A_F\\.
\eeq
An agent $A_U$ in the role of user of the data supplies queries $Q_U$, which may be accepted:
\beq
A_U &\promise{+Q_U}& A_T\\
A_T &\promise{-Q_T}& A_U
\eeq

If faithful to its inputs, the curator/receiver agent is now in possession of data and feature matching criteria,
which it can promise to three different downstream channel agents, 
\beq
R \promise{+X \;|\; (X_S\cap X_R) \;,\;(\phi_F\cap\phi_R) \;,\; C} A_Q\\
R \promise{+X \;|\; (X_S\cap X_R) \;,\;(\phi_F\cap\phi_R) \;,\; C} A_K\\
R \promise{+X \;|\; (X_S\cap X_R) \;,\;(\phi_F\cap\phi_R) \;,\; C} A_V.
\eeq
where $X$ is the final collation of data, classified into features.
Each of these recipients selects a projection $C$ of 
the data conditional on its inputs and a context decomposition $\{C\}$,
\beq
A_C &\promise{+ \{C\}}& A_Q,A_K,A_V\label{ctx}\\
A_Q,A_K,A_V &\promise{-C}& A_C.
\eeq
Creating these independent formal agents could be considered an artefact
of the batch processing implementation of agent analysis used in deep learning.
It is somewhat overkill for a more agile agent at the cognitive edge of a system.
The channels
\beq
A_Q \promise{+ Q(X,C) \,|\, X,C} A_T\\
A_K \promise{+ K(X,C) \,|\, X,C} A_T\\
A_V \promise{+ V(X,C) \,|\, X,C} A_T
\eeq
are recombined by an agent $A_T$ called the transformer:
\beq
A_T &\promise{-Q(X,C)}& A_Q\\
A_T &\promise{-K(X,C)}& A_K\\
A_T &\promise{-V(X,C)}& A_V
\eeq
which then promises to rank the matching data selections and return the root mean square of the values in
order of rank to the user agent $A_U$, from which the user typically selects the maximum:
\beq
A_T &\promise{+\left( \frac{K \cap Q}{\sqrt D} \right),V \;|\; K,Q,V} & A_U\\
A_U &\promise{-\max_V(f(\ldots))}& A_T.\label{att1}
\eeq
The rigmarole, expressed by this multitude of serialized processing agents, exposes both the detail and the potential
fragility of the selection of results as an optimization pipeline. Its complexity helps to explain
the large energy expended in undertaking the procedure for large data sets; but, equally, it shows
how that bulk process can be stripped down to minimal cognitive exchanges for simpler problems.
Unlike a matrix multiplication, every arrow may be considered an on-going continuous change process.

As written, in symbolic terms, there is no reference of the scales
(particularly the timescales) over which data flow around the promise
network. On general principles, the larger the data the longer the
time it takes to respond, thus large data models will be slow for two
reasons: statistical changes to trends and patterns are slow to be
accumulated and reach a level of significance, and the act of
collating them adds a second penalty. Thus big data, from a wide area
source, are slow data.  By contrast, individual agents that observe
specific phenomena, in isolated contexts, can respond much more
quickly and avoid the interference of competing signals. This leads to
a tradeoff, or a separation of concerns, between fast and slow
variables.

Lastly, we remark that this promise graph should not be confused with a knowledge graph. It has more in common with
a flowchart. However, we can now consider the representation
of knowledge, on the interior of the transformer agent $A_T$, and the role of the $\gamma(3,4)$
representation for $\phi_d$ in $A_F$.

\subsection{Knowledge graphs}

A knowledge graph is a virtual representation, in which each node in a
graph represents a `knowable'. A knowable might be as little as a word or as much as a whole book,
an image, or a sub-graph of component items.
We use nodes to represent the same `tokens' as a matrix $X$,
with attributes spanned by $\phi_d$ represented as links to other
nodes---though certain data tend to align better with
either graphs or vectors, rarely both. There is much freedom to make these choices, and various
opinions about how to perform this decomposition in different
applications.  Many of the perceived limitations of knowledge graphs
stem from overly limited perceptions of how they should be used,

Using a promise representation as an intermediate form of information
semantics, we can see that the structure of attentive information is
universal.  The useful purpose of the exercise is to see how we might
unify knowledge graph representations with machine learning in a
hybrid architecture. This is more subtle than simply defining graph
machine learning\cite{gl1,gl2} or trying to replace a graph with a vector
representation because `that's the hammer we know'. Some want to replace graphs with
vectorized learning, some want to use vectorized learning to build or
complete graphs. Both of these conversion therapies seem to ignore the specific
value of each other. The future more likely lies in a hybrid service approach.

The aim of a knowledge map or graph is to capture and sew together
intact samples of original experience, with intent and interpretation. By contrast, the aim of
a vectorized language model is to {\em replace} the original intention and
original meaning with a `strawman' as a generative framework in which one can shape {\em new}
stories stochastically within certain guiderails.  In traditional
semantic web graphs, $\phi_d$ are defined as an exercise in formal
typing called an
ontology\cite{ontologies,ontologydb,fanizzi1,fanizzi2}. This is not
easy to vectorize, since there is no beneficial link between types
and geometry except in special cases. However, in
the Semantic Spacetime $\gamma(3,4)$ representation,
which places $\phi_d$ into four major
classes\cite{burgess_sst2025} associated
with spacetime process, we simply embed elements of informed knowledge
within a contextualized process of retrieval.

Because much of the graph theoretic literature is about undirected graphs, 
most machine learning literature is also about undirected graphs, which is
an immediate handicap. An important feature of directed graphs is the existence
of dead-ends and blind alleys (so-called absorbing states), which
provide finite path lengths and definite outcomes\cite{burgess_sst2025}.

The semantic spacetime model argues that causal intent is the major
classifier in modelling spanning sets (just as the working week trumps
daily pattern in figure \ref{weekly}). Graph attributes may associated
with various real work processes: some static, such as security
modelling \cite{burgessC12}, some highly dynamic such as smart
environments and digital twins\cite{dtwin}.
\begin{figure}[ht]
\begin{center}
\includegraphics[width=6.5cm]{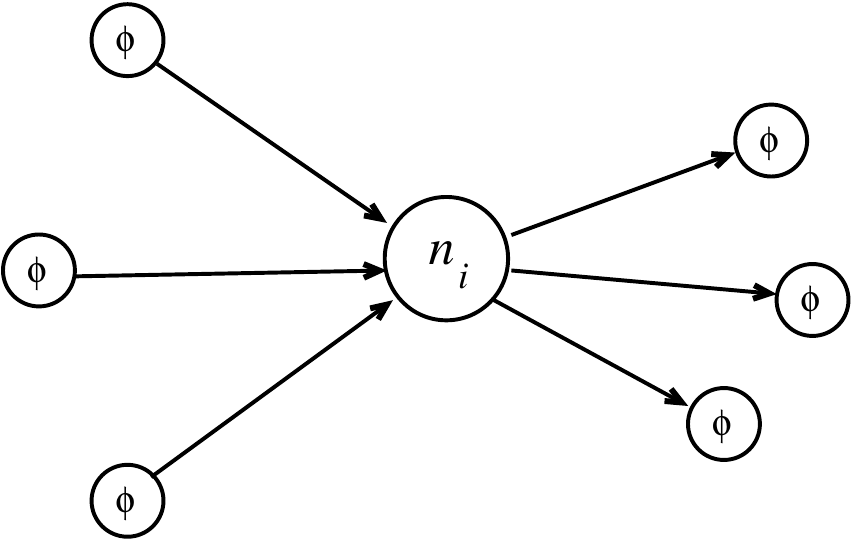}
\caption{\small A Semantic Spacetime graph structure. Each node is a knowable text and a unique
identifier for compression. Every attribute is a directed link to another node. For highly dynamic
information, one has to decide whether nodes are invariants or variables, which complicates the
semantics of links.\label{graph_gamma}}
\end{center}
\end{figure}
In a graph $\Gamma(N,L)$, the links may be based on the in-graph assessments $\alpha_R(\pi^+\pi^-)$, based
on source characteristics or observer assessments $\alpha_T(\pi^+\pi^-)$, for example:

\beq
L(\phi_d)  = \left \lbrace    
\begin{array}{ll}
S   & \text{Full node text.}\\
v[]  & \text{Text fractionation decomposition.}\\
L(\gamma(3,4))  & \text{A semantic spacetime type.}\\
\text{\sc evc}(A+A^{\rm T})  & \text{Eigenvector centrality seen by $T$}\\
\text{\sc btc}(A^{\rm T})  & \text{Betweenness centrality seen by $T$}\\
w(b_S \cap b_R)  & \text{The link weight, e.g. probability.}\\
\text{ctx} , \chi & \text{The selected context as a subset of the curated intended contexts.}
\end{array}
\right.
\eeq
The context sets `ctx' or $\chi$ are the representation of equation (\ref{ctx}).
These are the static references to be matched with a corresponding declaration
of context in a query.

Unlike a batch vector, in which each row $\vec x \in X$ is an
independent variable in which any graphical arrow between tokens is
only implicit in its $\phi_d$, in a knowledge graph representation,
the links between token nodes is explicitly encoded as a typed arrow.
concepts---they are linked intentionally.
In a knowledge graph, the $\phi_d$ are not agent-interior values (though
generalized graph databases can arrange this).  They are also nodes
$n_i$ in the graph. A graph relation is always an automorphism. This
is reflected in the widespread use of `self-attention' in machine
learning, as in Principal Component Analysis\cite{duda1}.

There are two kinds of graph curation:
\begin{itemize}
\item {\em Autonomously emergent graphs}: these are graphs in which each agent independently promises a data value, based
on its own internal processes. Each agent $A_i = \{ S, R, \ldots \}$ 
operates at its own rate, with its own time $t_i$, as assessed by the
external observer $T$.

\item {\em Curated graphs}: these are graphs that originate entirely from a single source $T$, which is the curator of the
graph. The information promised by each node originates from promises by $T$ and is merely keeping a storage promise,
which it may do with a certain fidelity to be discussed.
Many authors have proposed somehow using LLMs to curate knowledge. This feels like a misuse of the tools.
There is an ideological barrier here: LLMs don't actually know
anything by themselves so they can't be considered as suitable
generators of knowledge graphs. Knowledge is the documentation of
familiar experience, whereas language models mediate the generation of plausible language.
\end{itemize}

\begin{figure}[ht]
\begin{center}
\includegraphics[width=6.5cm]{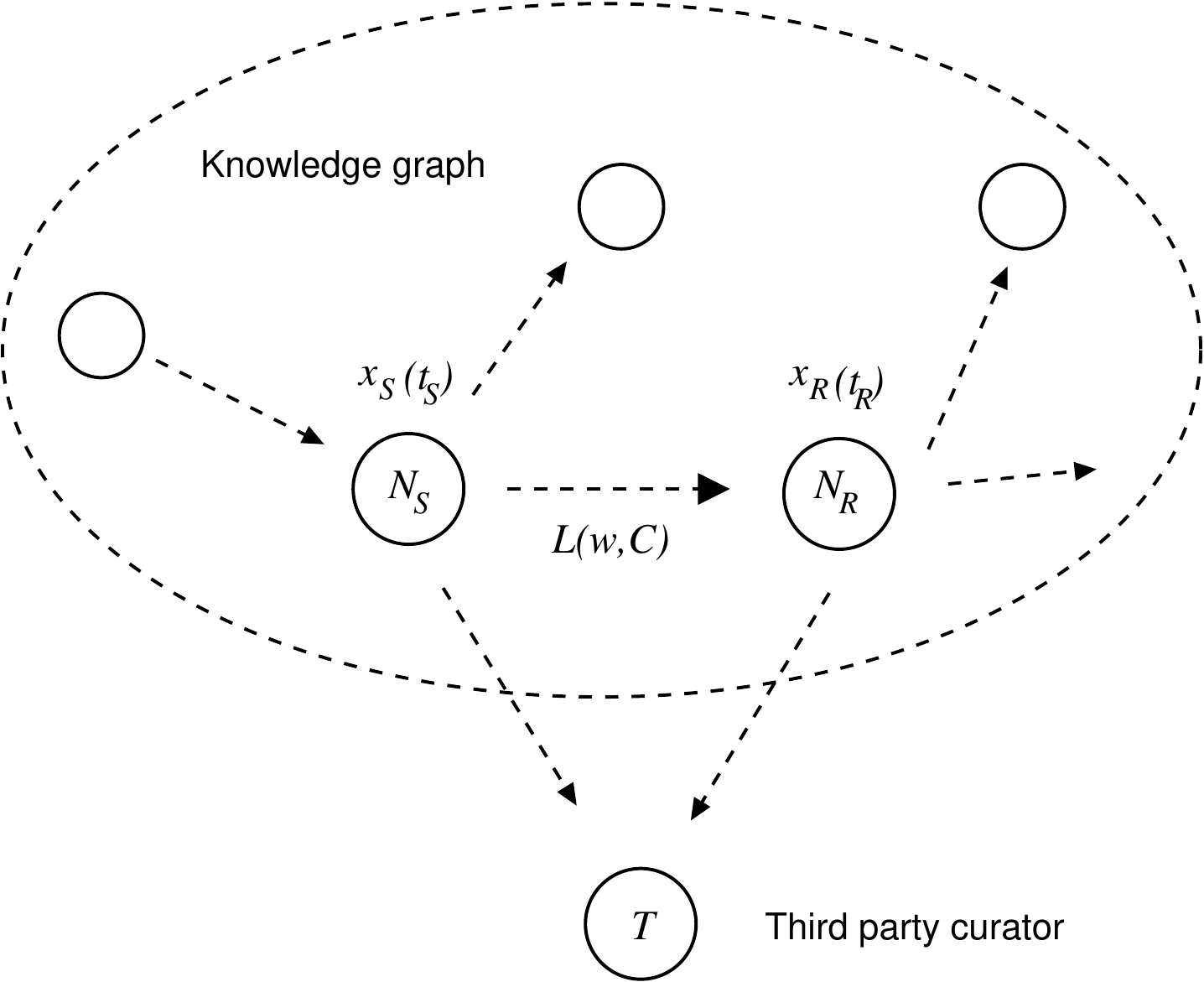}
\caption{\small In general, relationship in an SST knowledge graph is the result of a three agent cluster of
promises. We can denote the two of the nodes as `sender' $S$ and `receiver' $R$ for convenience,
though these roles will alternate. The most important invariant role is that of the third party
observer agent $T$, which is querying the information promised by the network nodes $N_S$ and $N_R$.\label{promise_graph}}
\end{center}
\end{figure}

See figure \ref{promise_graph}.
The assumption is that all graphs are modelled as promise theoretic
agents. Presently, in the world of information technology, knowledge
structures are typically built as `databases' and `knowledge graphs'
which are static and which are not maintained autonomously by each
node, but are assembled by a single command process that determines
all input and output in a centralized way. This industry standard also
applies to machine learning scenarios that are trained with static
snapshots of data collected. It should be considered a special case of
what we discuss here, and we do not begin by assuming that scenario.
In dynamic `smart environments', for instance, nodes in a graph may
represent different devices, equipped with sensors and processing
capabilities. These distributed software agents are involved in
independent data collection throughout a complex environment of
change.  That is the scenario we discuss here using Promise Theory's
axioms of independence as an arbiter in interpreting the consequences
for learning and representing `knowables'.

There are so many kinds of graph, with different assumptions, that
attempting to say something general is only plausible at this most
generic level. Many deal only with undirected graphs, however these
are very limited.  Directed and undirected graphs have very different
properties, which $\gamma(3,4)$ recognizes and accounts for.  Social
graphs are totally different from knowledge graphs, which are
different from electrical circuitry, and recommendation systems, and
so on.  Some graphs are simply sequences or stories, as paragraphs of
natural language.  Some graphs form maps used to trace an ongoing
scenario, like a complex group of bank transactions.  Task planning
for autonomous vehicles like robots and drones goes hand in hand with
this kind of causal intent.  The key point here is to show that the
attention mechanism, which separates context as an attribute like any
other recycled annotation, offers a simple way of utilizing data
classification.

\section{Summary and Conclusions}

By constructing a generalized model of `attention', we see how
monolithic batch learning and knowledge graphs can be combined.  The
principal elements of attention span the full process of learning and
querying, for an arbitrary representation. The approach is well suited
to distributed agent-centric systems, and can be parallelized across
agents provided one reduces the dependency of estimation on precise
relative normalizations. The larger and more monolithic we
make agents, the less responsive and less `intentional' or accurate
they are likely to be.

Using a promise representation, as an intermediate form, we see that
the structure is universal.  In bulk learning (e.g.  LLMs), the agents
are monolithic and consequently impractical to update in realtime. For
specialized agents (e.g.  CFEngine), learning in realtime is the norm
and can be performed on a much smaller scale to separate
specialization from generic capabilities.  This indeed suggests that
`attention is all you need'---but needn't be attached to neural
networks or huge vectorized data sets.

Humans think and reason by telling stories or sewing together
narratives that can later be rationalized by carefully choosing
arrows to join independent events and concepts together. Within a
knowledge graph, the interpolation of knowable elements forms an
intentional linguistic trajectory. This is a key part of knowledge
representation as stories, explanations, etc.  However, the focus on
linguistic legerdemain now at the forefront of LLMs, may ultimately only impede the
clarity of results by regenerative entropy. This is important when
reasoning is intended for autonomous machinery.
For example, we make use
of bullet points for a reason: too much padding may detract from the
content. The padding in natural language is not for information's sake 
but rather for human trust building. Graph
representations are more economical and consequently more precise on a
symbolic level.  Some stories can thus be represented as paths through
a graph, if there are sufficient nodes and links.  Intentional
knowledge can never be computed from generic probabilities without
some shadow representation.

While language models invest resources in outputting smooth human
language, knowledge graphs invest in representing precise causal
intent.  Stochastic language generation requires smooth data in order
to succeed, partly because language itself is a process which is
gradual in its unfolding. Language models will therefore never be able
to yield precise causal answers to precise questions, without
bypassing probability to merely paraphrase intentional knowledge.
Conversely, knowledge graphs will always produce somewhat stilted
output, measured against natural language. Thus language models are
preferred for the generation of long texts (the smaller the output,
the less sensible it is likely to be). For knowledge graphs the
situation is reversed. The quest for actual intelligence has to meld
these two aspects of reasoning together.

There is a sense, in current machine learning, that people are fixing
the graph to suit the method of vector processing---relying on the
brute force crunching of vectorized data and GPU computation, because that's
what they know. Although
one {\em can} always shoehorn a graph into a vector representation,
the results are unlikely to be conducive to meaningful algorithmic
processing. The natural
approach would therefore be to separate concerns, indeed separate
fast cognitive variables and slow batch processing of `big data' from one another
as independent services.

This touches on an unresolved aspect to both vectors and graphs: namely what
happens when data are highly dynamic. Bulk machine learning, which
dominates the present discourse, creates frozen snapshots of prior
experience.  Live action updates only enter through queries in these
models.  However, the definition in terms of promise agents and
realtime learning makes realtime updates plausible for smaller agents.
Conventional wisdom, from physics, suggests that a weak coupling
approach to modelling is the way to handle changing state: a
separation of scales between an approximately invariant background and
a fluctuating process relative to it.  Graphs are not well suited to
representing numerical data because the cardinality of numbers is
infinite. Graphs favour small domain-range sets.  However, it's also
true that the matrices $X$ are quite inflexible, since they require
uniform dimensionality. In fact they are equivalent to simple graphs
$\text{rows}\times\text{columns} = \text{LT}(\gamma(x,1)) \times
\text{EP}(\gamma(x,3))$.  The shift from batch snapshot processing to
continuous cognitive updating points to a refactoring of the AI
infrastructure model in terms of specialized agents, like CFEngine,
will facilitate the real embedding of smart adaptation.  This is
closely analogous to to the way that brain cells are specialized to
perform their tasks as part of a continuous pipeline. An architectural
shift is the only practical way to handle real world sensory
adaptation while benefitting from bulk knowable information.

Language models have beguiled many, because smooth language generates
a sense of false familiarity. However, language generation has damning
problems that make it unsatisfactory for obtaining predictable
results---a key requirement in safety situations, such as autonomous
vehicles, emergency services, defence, etc. Intent has to come to the
rescue, e.g. by invoking the predictability of fixed absorbing
states\cite{burgessC11}.
Even for more harmless best-effort explanations, we do not really want to generate
stories probabilistically: we want actual facts in a story to have
been built intentionally. This is where Knowledge Graphs potentially
enter the picture as a representation of actual `causally connected
phenomena'.

Language models also generate other quasi-symbolic patterns, by
predicting the next `move' or symbol in a sequence, e.g. in bioinformatics.  There is clearly
a role for their kind of creative speculation.  During this matching
process $Q\cap K$, an atomization of the elements of the training
data, something like a DNA spectral analysis, denatures embedded
intentionality\cite{burgess_intent}. This is what we must recover. In
the SSTorytime software, this is applied both as a data vector
(postgres \verb+ts_vector+) and as an n-gram fractionation in {\tt
  text2N4L}\cite{SSTorytime} but in different ways. Vectors are good
for searching, while graphs are good for intentionality.

Both knowledge graphs and vectorized transformers align when
the third party curation of the data has the same underlying intent as the
querying agent. This tells us that we should retain causal information somehow.
Intent is the key---and it is this problem which is
yet to be explored in detail, perhaps using graphs within the query language too.

\bibliographystyle{unsrt} % biber
\bibliography{spacetime,bib}

\appendix
\section{Pseudo-periodic learning over time-series phenomena}

Time-series data are particularly relevant for cognition in all kinds of agents.
The ability to form an expectation of future patterns depends on what kind of model
one uses to minimize noisy variations.
The recent behaviour of a computer can be summarized by non-Markovian
processes, during periods of change, and by hidden Markov models
during steady state behaviour, but one still requires a
parameterization for data points. Such models must be formulated on a
periodic background\cite{burgessIJMPC}, owing the importance of
periodic behaviour of users. The precise algorithm for averaging and
local coarse-graining is somewhat subtle, and involves naturally
orthogonal time dimensions which are extracted from the coding of the
database. It is discussed here using an ergodic principle: a
bi-dimensional smoothing is implemented, allowing twice the support
normally possible for the average, given a number of data points. This
provides good security against ``false positive'' anomalies and other
noise.

Consider a pseudo-periodic function, with pseudo-period $P$,
\beq
q(t) &=& \sum_{n=0}^\infty\; q(nP+\tau)\qquad\qquad (0 \le \tau < P)\nonumber\\
&=& \sum_{n=0}^\infty\;\chi_n(\tau).
\eeq
The time coordinate $\tau$ lives on the circular dimension. In
practice, it is measured in $p$ discrete time-intervals $\tau =
\{\tau_1,\tau_2,\ldots \tau-p\}$.  In this decomposition, time is a
two-dimensional quantity. There are thus two kinds of sliding average
which can be computed: average over corresponding times in different
periods (topological average $\langle \chi(\tau)\rangle_T$), and
average of neighbouring times in a single period (local average
$\langle\chi(\tau)\rangle_P$):
\beq
\langle \chi(\tau)\rangle_T &\equiv& \frac{1}{T}\; \sum_{n=l}^{l+T} \chi_n(\tau)\nonumber\\
\langle \chi(n)\rangle_P &\equiv& \frac{1}{P}\; \sum_{\ell=\tau}^{\tau+P} \chi_n(\ell)
\eeq
where $P,T$ are integer intervals for the averages, in the two time-like directions.
Within each interval that defines an average, there is a corresponding definition of the
variation and standard deviation, at a point $\tau$:
\beq
\sigma_T(\tau) &\equiv& \sqrt{ \frac{1}{T}\; \sum_{n=l}^{n=l+T} \left(\chi_n(\tau)-\langle \chi(\tau)\rangle_T\right)^2  }\nonumber\\
\sigma_P(n) &\equiv& \sqrt{\frac{1}{P}\; \sum_{\ell=\tau}^{\ell=\tau+P} \left(\chi_n(\ell)-\langle \chi(\ell)\rangle_P\right)^2 }.
\eeq
Sliding window versions of these may also be defined, straightforwardly from the preceding
section:
\beq
\overline\chi(n)_P &\equiv& (\chi\,|\,\overline\chi(n)_P)\nonumber\\
\overline\chi(\tau)_T &\equiv& (\chi\,|\,\overline\chi(\tau)_T).
\eeq
Here one simply replaces the evenly weighted sum, with an iteratively
weighted sum.  The difference between the average and the current
value is expressed by the delta symbol $\delta_T\chi = \chi-\langle\chi\rangle_T$ etc.

Using these averages and deviation criteria, we have a two-dimensional
criterion for normalcy, which serves as a control at two different
time-scales. One thus defines normal behaviour as
\beq
\{\delta_T\chi(\tau),\delta_P\chi(n)\} < \{2\sigma_T(\tau),2\sigma_P(n)\}.
\eeq
These may be simply expressed in geometrical, dimensionless form
\beq
\Delta(\tau,n) = \sqrt{\left(\frac{\delta_T\chi(\tau)}{\sigma_T(\tau)}\right)^2+\left(\frac{\delta_P\chi(n)}{\sigma_P(n)}\right)^2},
\eeq
and we may classify the deviations accordingly into concentric, elliptical regions:
\beq
\delta(\tau,n) < \left\{
\begin{array}{c}
\sqrt 2\\
2\sqrt 2\\
3\sqrt 2
\end{array}
\right.,
\eeq
for all $\tau,n$.
which indicate the severity of the deviation, in this parameterization.
This is the form used by cfengine's environment engine.

\begin{figure}[ht]
\begin{center}
\includegraphics[width=7.5cm]{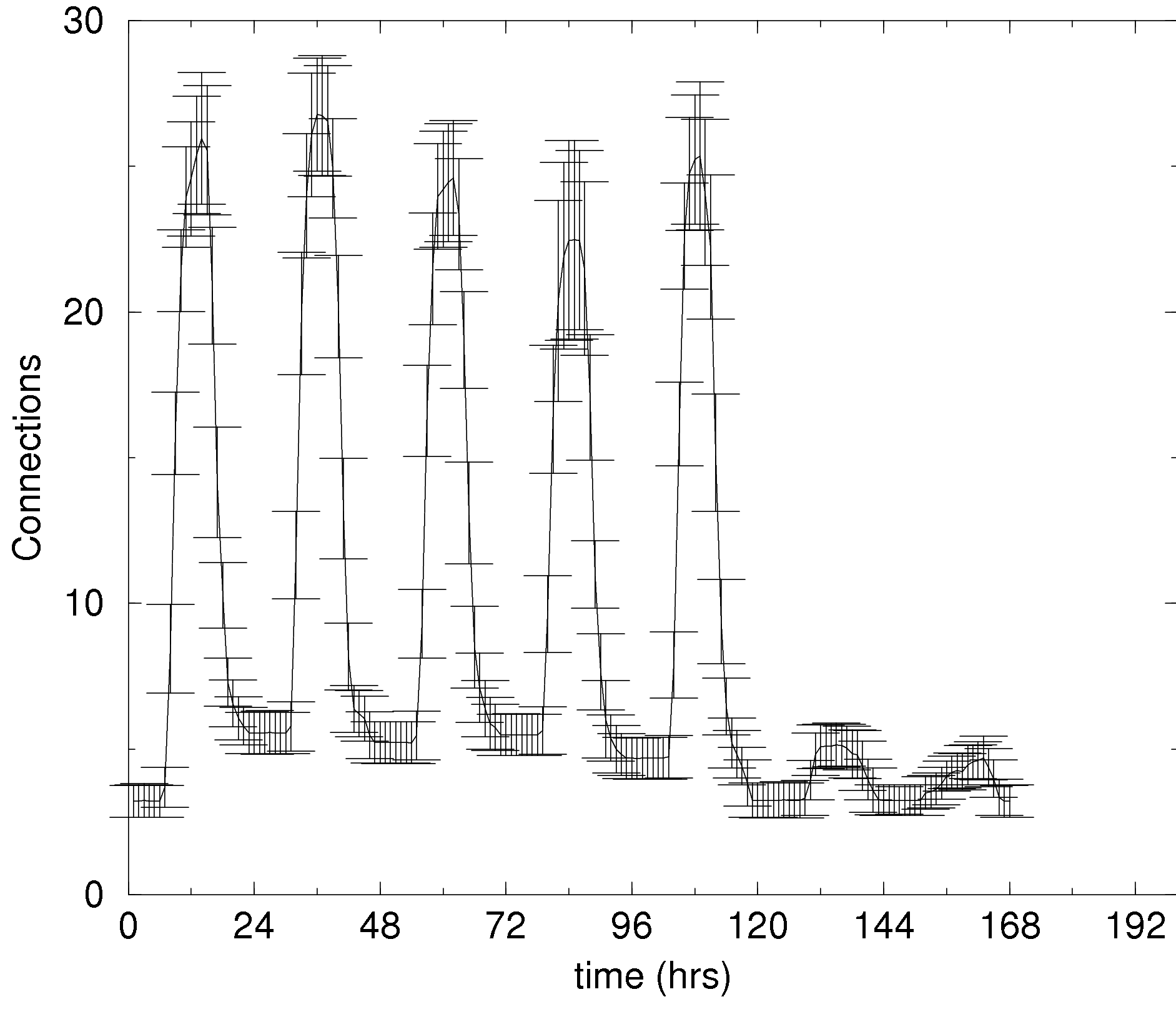}
\includegraphics[width=7.5cm]{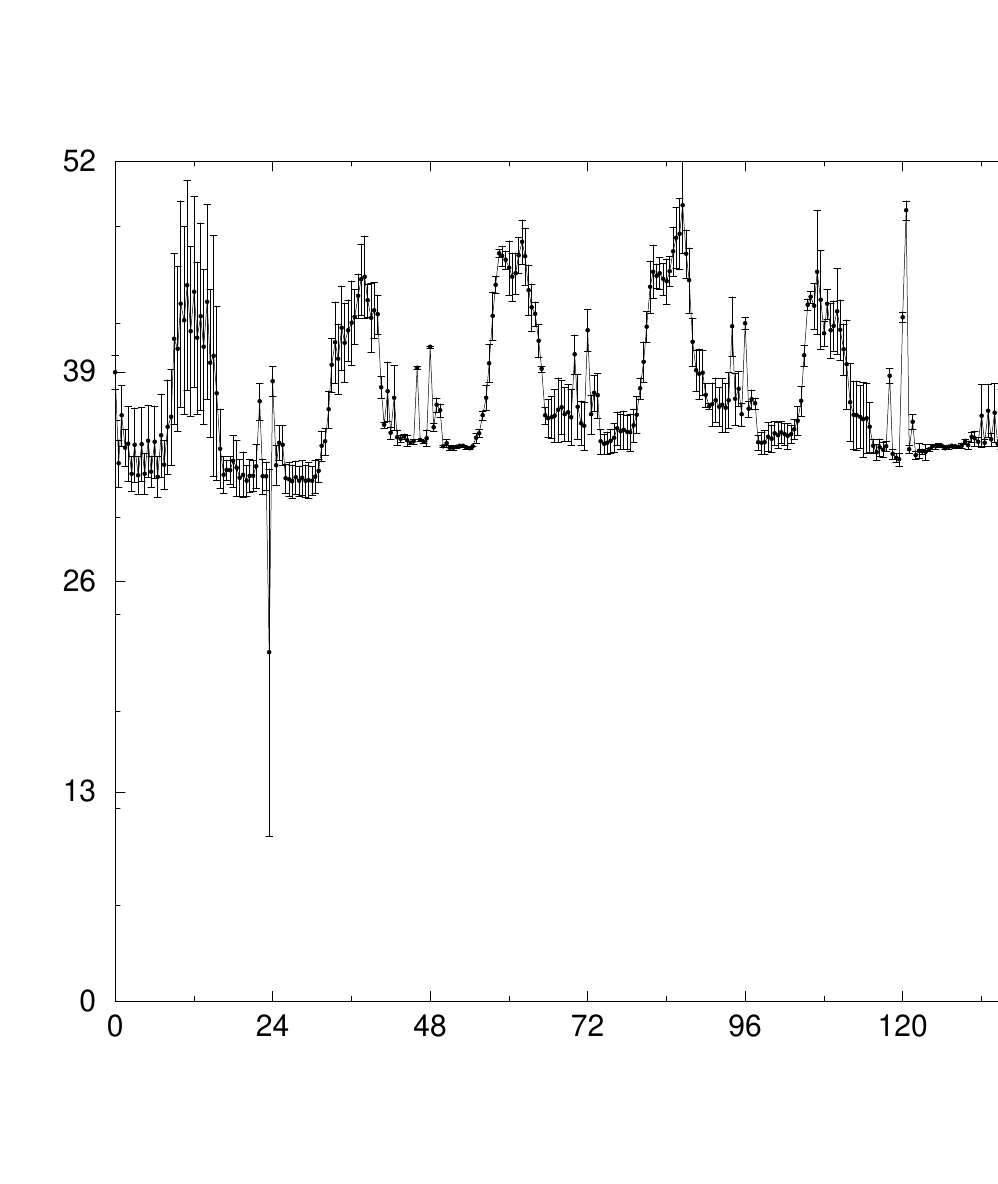}
\includegraphics[width=7.5cm]{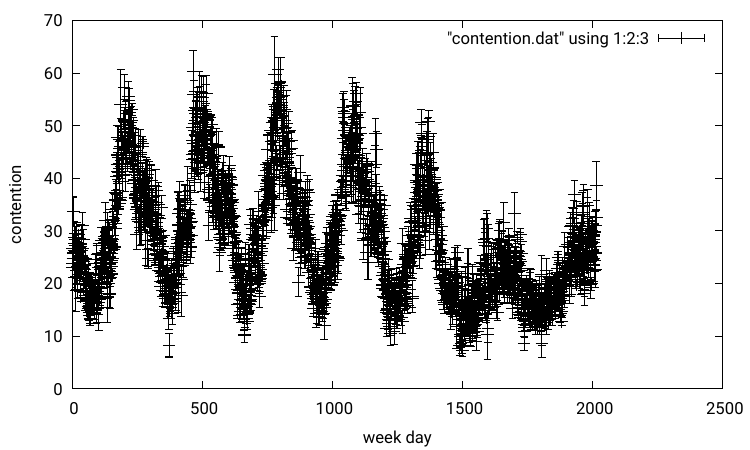}
\caption{\small Three randomly chosen data observables taken from a
  network server from \cite{burgessC8}.  We see that the human working week is imprinted on
  both as peaks for each day, smaller at the weekends, and the
  averages over many weeks yield error bars of varying
  size. Each periodically identified time has an activity or energy level
associated with it, also apparent in the variable width of the error bars.
Note also the presence of an anomaly, identifiable by deviation from neat self-similarity.\label{weekly}}
\end{center}
\end{figure}

\end{document}